\documentclass{article}

\usepackage{microtype}
\usepackage{graphicx}
\usepackage{subfigure}
\usepackage{booktabs}
\usepackage{enumitem}
\usepackage{dialogue}

\usepackage{hyperref}

 \usepackage[accepted]{icml2024}

\icmltitlerunning{ShodhLab}

\begin{document}

\twocolumn[
\icmltitle{Can LLMs Compute with Reasons?}

\begin{icmlauthorlist}
\icmlauthor{Harshit Sandilya}{lab,mnit}
\icmlauthor{Peehu Raj}{inferq}
\icmlauthor{Jainit Sushil Bafna}{lab,iiith}
\icmlauthor{Srija Mukhopadhyay}{lab,iiith}
\icmlauthor{Shivansh Sharma}{sgvu}
\icmlauthor{Ellwil Sharma}{lab}
\icmlauthor{Arastu Sharma}{lab}
\icmlauthor{Neeta Trivedi}{inferq}
\icmlauthor{Manish Shrivastava}{iiith}
\icmlauthor{Rajesh Kumar}{mnit}
\end{icmlauthorlist}

\icmlaffiliation{lab}{ShodhLab.ai}
\icmlaffiliation{mnit}{Raman Lab, Malaviya National Institute of Technology, India}
\icmlaffiliation{iiith}{Language Technologies Research Center, KCIS, IIIT Hyderabad}
\icmlaffiliation{inferq}{Inferigence Quotient Private Limited}
\icmlaffiliation{sgvu}{Suresh Gyan Vihar University, India}

\icmlcorrespondingauthor{Harshit Sandilya}{harshitsandilya@gmail.com}

\icmlkeywords{Large Language Model, Reasoning of LLM, Mathematics LLM}

\vskip 0.3in
]

\printAffiliationsAndNotice{}

\begin{abstract}
Large language models (LLMs) often struggle with complex mathematical tasks, prone to "hallucinating" incorrect answers due to their reliance on statistical patterns. This limitation is further amplified in average Small LangSLMs with limited context and training data. To address this challenge, we propose an "Inductive Learning" approach utilizing a distributed network of SLMs. This network leverages error-based learning and hint incorporation to refine the reasoning capabilities of SLMs. Our goal is to provide a framework that empowers SLMs to approach the level of logic-based applications achieved by high-parameter models, potentially benefiting any language model. Ultimately, this novel concept paves the way for bridging the logical gap between humans and LLMs across various fields.
\end{abstract}

\section{Introduction}
\label{introduction}
Large Language Models(LLMs) have proved extremely capable of performing various generative tasks \cite{brown, chen-a}. Open source models have also shown considerable success in this field \cite{open-llm-leaderboard}. However, reasoning tasks, especially mathematical reasoning, continue to pose a massive challenge for all of these models \cite{hendrycks, espejel}. 

There has been considerable work on improving the performance of LLMs for mathematical reasoning tasks, including training on specialized datasets through different fine-tuning approaches \cite{yuan, luo} or continual pre-training \cite{taylor}. Other methods include using different prompting mechanisms to obtain better answers and enhance the reasoning capabilities of these LLMs \cite{wei, chen, gao}.

Additionally, new methods that focus on creating smaller models trained on "textbook-quality" data have also seen tremendous success. \cite{phi1.5}.

\section{Related Work}
\label{related}
\subsection{Improving Mathematical Reasoning of LLMs}
Various methods are employed to improve the mathematical reasoning capabilities of LLMs. One commonly applied method is continual pre-training \cite{azerbayev}. The model is trained on large-scale mathematical datasets, fine-tuning it by continuing the pre-training process. 

Another approach employed is supervised fine-tuning \cite{yuan, luo}, where high-quality question-answer pairs are collected through various techniques and then used to fine-tune the model to enhance its performance. Synthetically constructed datasets are often used for this process. These methods use LLMs to generate the data, followed by various augmentation methods \cite{yuan, yu, li} to filter the data. 

Various reasoning frameworks are also implemented to bring out the best reasoning answers. This includes prompting-based \cite{wei, chen} and self-consistency \cite{wang} methods where the model uses majority voting to decide among various rational paths. 

The most commonly used approaches are Chain-of-Thought reasoning \cite{wei, nye} and Program-of-Thought reasoning \cite{chen}. Over the recent past, new methods like Equation-of-Thought Distillation \cite{xunya}, which work on a similar principle, have also emerged. 

There is also the emergence of Program-aided Language Model (PAL) \cite{gao}, where code is generated from LLMs, which is later passed through an external API to generate the final output. A similar approach also uses symbolic solvers to implement the same \cite{heyueya}.

\subsection{Knowledge Distillation based methods}
Knowledge Distillation \cite{hinton, magister, shridhar} is a method to transfer knowledge from a general usage LLM to a smaller, efficient Small Language Model (SLM) without losing validity. 

Various methodologies might be used for the same, which include response-based knowledge transfer \cite{hinton, jin}, feature-based knowledge transfer \cite{roheda, heo}, and relation-based knowledge transfer \cite{yim, li}. 

Reasoning generated from prompts like Chain-of-Thought (CoT) \cite{wei}, Program-of-Thought (PoT) \cite{chen}, and Equation-of-Thought (EoT) \cite{xunya}, along with their combinations, might also be used to distill knowledge from LLMs to SLMs.

\begin{figure*}
  \centering
  \includegraphics[width = \textwidth]{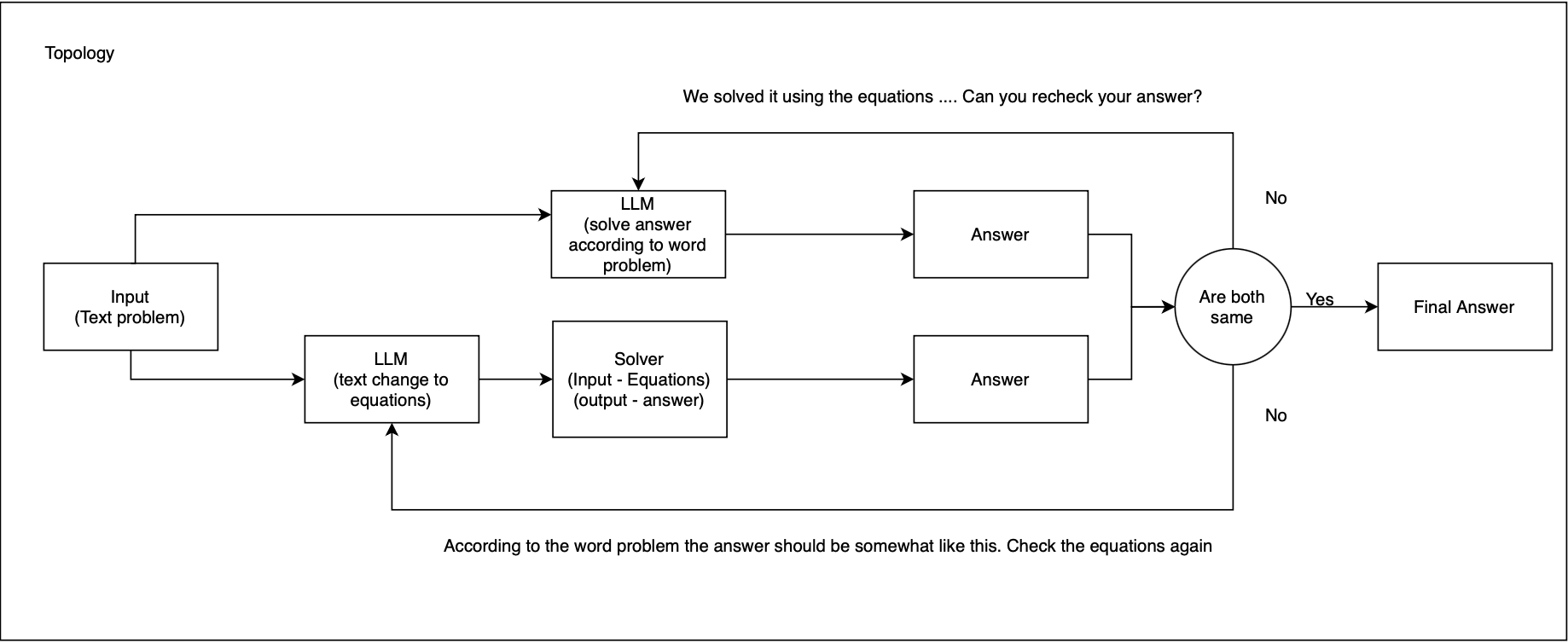}
  \caption{Network Topology}
  \label{fig:toloplogy}
\end{figure*}

\subsection{Ensemble Methods}
Ensemble methods \cite{ganaie} involve creating multiple models and combining them to produce better results. Various techniques combine these models, such as majority voting, confidence scoring, and aggregation. Additionally, other models can be used, as demonstrated through LLM-Blender \cite{jiang}. 

Our approach is closely related to the work done in TinyGSM \cite{tinygsm}, which studies the importance of high quality datasets for enhancing the mathematical reasoning capabilities of SLMs. Another closely related work is MathPrompter \cite{imani} which uses Zero-shot chain-of-thought prompting to generate multiple reasonings for the same function to raise the confidence level in output results.

However, we improve on these by using a distributed network of N model pairs. These models perform computations in parallel and contribute to the voting process. Thus, we increase efficiency while being able to obtain more accurate results.

\section{Methodology}
\label{methodology}
We introduce the "Inductive Learning" approach for reasoning enhancement in this section. As seen before, many advancements have used solvers to rectify the error. The main area of improvement seen was the missing reasoning. Our network improves upon the reasoning by introducing a second LLM, and to keep the probability of correctness still equivalent to the single LLM-based reasoning, we made some modifications to the network.

\subsection{Reasoning Ability}
The network, as shown in Figure~\ref{fig:toloplogy}, is the basis of the reason. The upper LLM is termed GP in the discussion below, while the lower one is termed EQ.

Our architecture leverages the context mechanism to auto-correct itself using hints from the other counterpart. Both the LLMs are used to solve a question. GP solves the question using logic and reasoning, while EQ converts the question to a computation task that can be accomplished using any programming language. Both produce results independent of each other, and finally, we compare the results to check them.

Here, two possibilities emerge:
\begin{enumerate}[label=\roman*,leftmargin=15pt]
  \item Both give the same answer: The answer might be correct. Forward it as output.
  \item Both give a different answer: One or both LLMs made a mistake. Give them hints in a context like 'The equations we used were' and 'Logically thinking we might get the answer.'
\end{enumerate}

Suppose the probability that a GP gives the correct answer is $P_{reason}$ while for an EQ, it's $P_{numerical}$. So, for the final output to match, we have a probability of $P_{total}=P_{reason}\times P_{numerical}$, which is reduced to the original probability as $P<1$. But the surety of getting the correct answer increases.

\subsection{Cross Inference}

\begin{algorithm}[tb]
   \caption{Cross Inference}
   \label{alg:inference}
\begin{algorithmic}
   \STATE {\bfseries Function} match(GP, EQ)
\STATE {\bfseries Input:} lists GP and EQ
\STATE {\bfseries Output:} index of the element in GP with the highest matching score

\STATE Initialize matching\_score $\gets$ a list of 0s with length equal to the length of EQ

\FOR{$i = 1$ {\bfseries to} length of GP}
  \FOR{$j = 1$ {\bfseries to} length of EQ}
    \IF{GP[i] = EQ[j]}
      \STATE matching\_score[j] $\gets$ matching\_score[j] + 1
    \ENDIF
  \ENDFOR
\ENDFOR

\STATE {\bfseries return} index of the maximum value in matching\_score
\end{algorithmic}
\end{algorithm}

\begin{figure}
  \centering
  \includegraphics[width=0.4\textwidth]{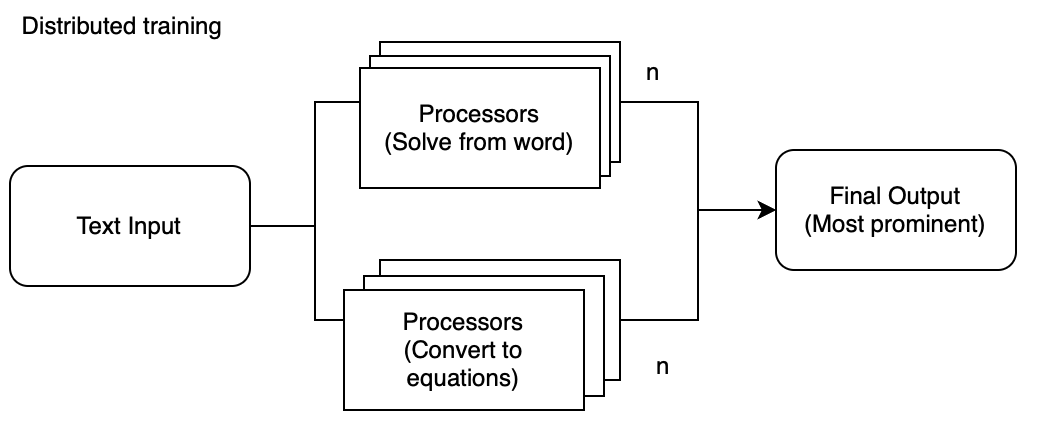}
  \caption{Distributed computation}
  \label{fig:parallel}
\end{figure}

To overcome this and get a meaningful result, we run the network as shown in Figure~\ref{fig:parallel}

\begin{figure}
  \centering
  \includegraphics[width=0.2\textwidth]{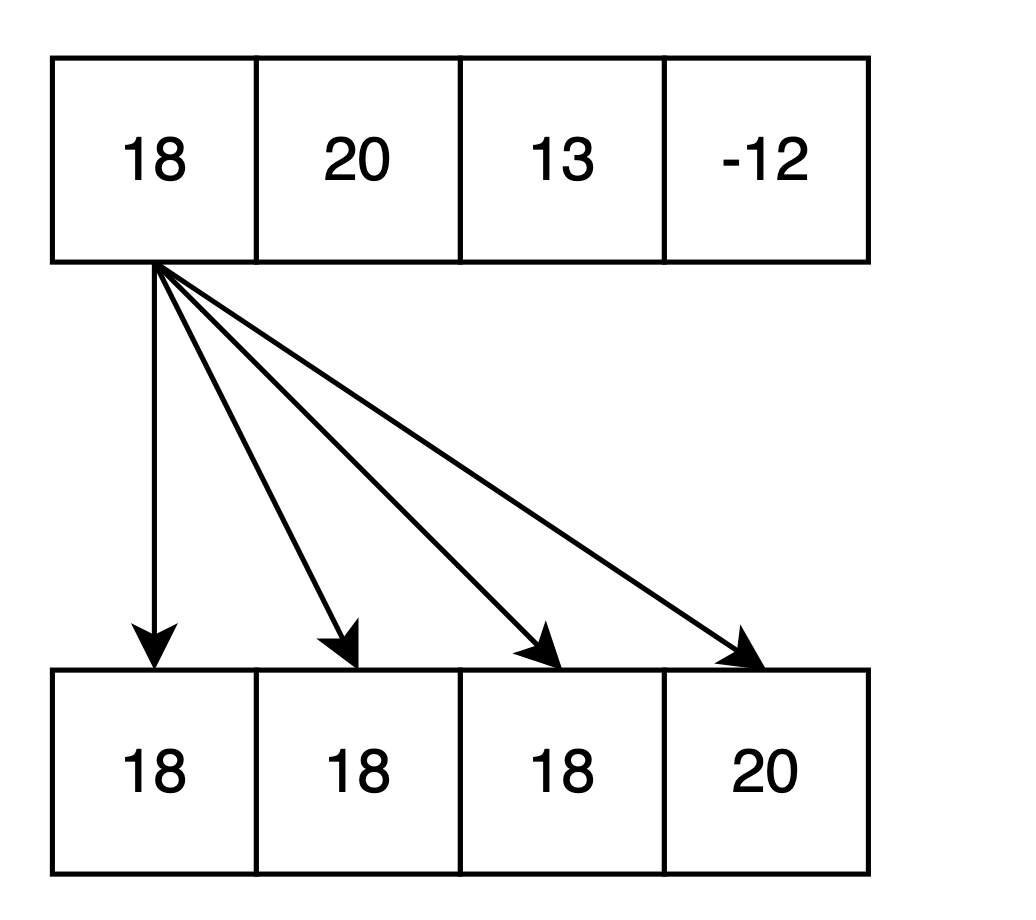}
  \caption{The cross inference for increasing probability}
  \label{fig:inference}
\end{figure}

Here, we have $n$ LLMs working in parallel and have to find the correct answer. We can use the following equation to find the probability of getting the correct answer:
\begin{equation}
    P_{total} = \sum_{k=1}^n P_{single}
\end{equation}
where \begin{equation}
    P_{single} = P_{reason}\times P_{numerical}
\end{equation}
Every pair here is mutually exclusive and learns only from its counterpart. We can modify the network to include cross-inference, increasing our probability of finding the correct answer. To do so, let's assume we store every LLM's output in a matrix. Thus, we have two such matrices; one contains answers for GPs, and the other contains answers for the EQs. We try to find the matching pairs that give the same output, as shown in Figure~\ref{fig:inference} 

We calculate a $matching\,score$ for every GP. And finally, the answer is the one with the highest matching score. Thinking in terms of Algorithm~\ref{alg:inference}, we can conclude:
\begin{equation}
    matching\,score_{i} = \sum_{j}^n \delta_{ij}
\end{equation}
Where $\delta_{ij}$ is zero if the pair doesn't match and one if it matches, and the probability that each $\delta_{ij}$ is correct depends on the independent probability of both being correct, giving us $P_{reason}\times P_{numerical}$ again. We then do the $max(matching\,score_{i})$ to find the final answer and send it forward.

Here, we assume that the network still learns from its counterpart, but instead, we now treat the inference over the whole network rather than treating them as individual networks. This gives us a boost over the conventional distributed architecture and increases our chances of finding the correct answer while still giving us an answer if no answer matches. This shows the potential to replace single LLM over logical tasks.

\section{Results}
\label{results}
\subsection{Experimental Setup}

\subsubsection{Dataset Description}

We utilized two distinct datasets, both containing questions from the Grade School Math 8K (GSM8K) dataset \cite{gsm8k}. These datasets, however are different in the nature of their answers. The first dataset, termed the "GSM8K-code dataset," contained answers as equations in the form of executable Python code snippets. Notably, the GSM8K-code dataset was generated by us utilizing the state-of-the-art GPT 3.5 model \cite{gpt} developed by OpenAI. The GSM8K-code dataset was utilized for fine-tuning the Python Equation (EQ) models responsible for generating answers as equations in the form of Python code. The second dataset, referred to as the "GSM8K-base dataset," is the same as the original GSM8K dataset on the HuggingFace dataset hub. GSM8K-base dataset was utilized for fine-tuning General Purpose (GP) models responsible for generating answers in natural language.

\subsubsection{Fine-tuning Models}

We employed an experimental setup consisting of eight Phi 1.5 models \cite{phi1.5}, fine-tuned utilizing a cluster of eight Graphical Processing Units (GPUs). Within this setup, four GPUs were allocated for fine-tuning GP models targeting the generation of general answers on the GSM8K-base dataset. Concurrently, the remaining four GPUs were dedicated to fine-tuning EQ models on the GSM8K-code dataset to facilitate the generation of Python code using the Hugging Face transformers API \cite{huggingface-api}.

\subsubsection{Inference}

Each GPU in the configuration was an NVIDIA A100, featuring 40GB of VRAM. To distribute the workload efficiently, four GP models and four EQ models were instantiated, with each GP model and EQ model assigned to separate GPUs. We employed the \texttt{multiprocessing} module of PyTorch \cite{pytorch} to spawn individual GP models and EQ models on their respective GPUs.

Upon completion of inference for each model, Torch's \texttt{torch.barrier} was utilized to synchronize all GPUs, ensuring consistent progress across the distributed system. Post-inference, the outputs of the EQ models were compared against the outputs of their corresponding GP models. In cases where disparities arose between the outputs, re-prompting and re-inference were executed exclusively for GP models whose outputs did not align with their corresponding EQ model outputs.

To validate the final outputs comprehensively, a voting mechanism was implemented, cross-verifying the outputs obtained from multiple models.

We can see varied results, from the actual increase in logic for solving math problems to some being missed at the last steps. We even see some wild cases where both are wrong and still give the same answer. We cover everyone in their different sections. To assess the overall efficacy of our network, we compared its performance with the benchmark set by the pre-trained Phi 1.5 model on the GSM8K benchmark. The Phi 1.5 model achieved a benchmark performance of 12.58\%, providing a reference point for our network's performance in algebraic reasoning tasks.

\subsection{GP Evaluation}
We initiated our evaluation by fine-tuning the GP (top LLM) on the GSM8K Train Dataset using the pre-trained Phi 1.5 model. Four runs of inference were performed on the GSM8K benchmark with the same dataset, considering four different pairs of LLMs in our network topology. The individual benchmark performances of the GP models (denoted as $GP\,I, GP\,II, GP\,III,and\,GP\,IV$) are shown in Table~\ref{tab:gp_output}

\begin{table}[h]
    \centering
    \begin{tabular}{cccc}
        \hline
        GP I & GP II & GP III & GP IV \\
        \hline
        33.49 & 32.46 & 34.44 & 34.72 \\
        \hline
    \end{tabular}
    \caption{Output for GPs}
    \label{tab:gp_output}
\end{table}

Slight variations in performance among the models are attributed to the inherent variability in the fine-tuning process. On average, the fine-tuned GP model demonstrated a performance of 33.1\%.

\subsection{EQ Evaluation}

The EQ (lower LLM) played a crucial role in generating equations, subsequently processed by a Python solver. Synthetically generated fine-tuning data was used to evaluate the EQ models (denoted as $EQ\,I, EQ\,II, EQ\,III\,and\,\,EQ\,IV$) on their ability to produce correct solutions. The individual performances of these models are as shown in Table~\ref{tab:eq_output}

\begin{table}[h]
    \centering
    \begin{tabular}{cccc}
        \hline
        EQ I & EQ II & EQ III & EQ IV \\
        \hline
        51.53 & 51.74 & 52.61 & 53.30 \\
        \hline
    \end{tabular}
    \caption{Output for EQs}
    \label{tab:eq_output}
\end{table}

On average, the EQ models demonstrated an effective performance of 52.30\%.

\subsection{Single Network Evaluation}
Running every network independently we can generate an output as shown in Table~\ref{tab:hallucination}. We could also see the edge cases here where both match and still couldn't match the actual answer.
\begin{table}[h]
    \centering
    \begin{tabular}{ccccc}
        \hline
        \ & I & II & III & IV \\
        \hline
        $GM=EQ \neq Ans$ & 27.81 & 27.85 & 29.04 & 30.78\\
        \hline
        $GM=EQ=Ans$ & 25.53 & 25.47 & 26.67 & 27.55\\
        \hline
    \end{tabular}
    \caption{Running Network Independently}
    \label{tab:hallucination}
\end{table}

These percentages represent the extent to which the outputs of the GP and EQ LLMs align during the cross-checking loops. The variations in matching percentages highlight the dynamic nature of the iterative improvement process, with increasing alignment observed across subsequent pairs.

Although the hallucinations are less, they greatly impact our results. Still, our network manages to reduce them with cross-inference. A single network can be seen decreasing in probability as assumed in the section before hence the actual run consisted of the cross inference.

\subsection{Topology output}
The network integration involves four loops of the LLM pair (GP and EQ), designed to cross-check and refine the generated answers iteratively.

When the entire network is run, the main output, represented by the Topology, is observed to be 50.29\%. This significant improvement is noteworthy when compared to individual components. Specifically, the improvement is as follows:

GP (top LLM) alone achieved around 33\%; with topology, we see an improvement of 17.3\% average. 
GP + EQ (matching and correct alignment) together yielded only 26\%, while the topology implementation was improved by 25.29\% on average.

The main objective was teaching the GP the logic behind the problem by breaking it down into equations close to what humans do in the real world. The concept of breaking the problem logically helped the LLM build its own reason and improve using the established facts.

\subsection{Comparative Analysis with other approaches}
In this section, we aim to provide a comparative analysis of our approach with other similar efforts in the field of enhancing mathematical reasoning using large language models. We will specifically discuss two notable works, one mentioned in the Phi 1.5 technical report and another in the context of TinyGSM.

\subsubsection{Comparison with Phi 1.5 Technical Report}
The Phi 1.5 technical report \cite{phi1.5} discusses a similar effort where Python output is employed through a Python solver for fine-tuning purposes, and the model's performance is evaluated on the GSM8K benchmark. In that case, the GSM8K performance of Phi 1.5 is reported to be 40\%.

In contrast, our novel network architecture, incorporating GP and EQ LLMs with synthetic data fine-tuning, achieves a main output of 50.29\%. This suggests a substantial improvement over the baseline set by the Phi 1.5 model. The utilization of cross-checking loops and the distributed learning method, particularly the LLM pair voting mechanism, contributes to the enhanced verification of overall output.

\subsubsection{Comparison with TinyGSM}
Another relevant work, discussed in TinyGSM \cite{tinygsm}, involves a detailed exploration of improving output through fine-tuning and verification loops with two different LLMs. While the output on GSM8K is reported to be better in this method, we argue that our approach excels in transferring logic more effectively.

\subsubsection{Comparison with TinyGSM}
In TinyGSM, a substantial increase in performance on the GSM8K test set was reported, rising from 44.6\% to 68.2\%. This represents an impressive percentage increase of approximately 23.6\%.

Comparative Analysis of Percentage Increases
Now, let's compare this with the percentage increases achieved in our approach:

\paragraph{Improvement in GP model:}

Our GP (top LLM) performance was initially around 33\%, and after the implementation of our network architecture, the GP improved by an average of 17.3\%.

\paragraph{Topology Output Improvement:}

The collaborative efforts of GP and EQ LLMs resulted in a notable improvement, with GP and EQ matched with ground truth is increasing by an average of 25.29\%.

Our network's distributed learning method, incorporating the collaborative efforts of GP and EQ LLMs in cross-checking loops, aims to transfer reasoning and logic in a more refined manner. The voting mechanism employed in our LLM pair contributes to a better-verified overall output.

While the \textbf{percentage increase in GP is substantial at 17.3\%}, it is important to note that the comparison metric in TinyGSM represents a different aspect of performance on the GSM8K test set. Additionally, \textbf{the combined improvement of by 25.29\%} further underscores the effectiveness of our network architecture.

Our approach demonstrates competitive percentage increases in relevant metrics, showcasing advancements in mathematical reasoning capabilities. The comparison with TinyGSM highlights the diverse methodologies employed in the field and the need for nuanced evaluation to understand the strengths and limitations of each approach.

\subsubsection{Future Research Considerations}
While our approach demonstrates promising results, we acknowledge the need for further research to comprehensively compare different methodologies. Comparative studies should delve into the intricacies of the approaches, considering not only benchmark performance but also the efficiency in logic transfer and verification mechanisms.

In conclusion, our network architecture presents advancements in mathematical reasoning capabilities, outperforming the benchmark set by the Phi 1.5 model. The collaborative efforts of GP and EQ LLMs, coupled with the distributed learning method, set our approach apart. Ongoing research endeavors will be crucial for a deeper understanding and comparison of different methodologies in this evolving domain.

\subsection{Logic Improvement}
The back loop hints to both EQ and GP to rethink their strategies. In due time, we can observe the growth of logic in the GP as it tries to get towards the answer. It is evident from the following examples we observed:

\begin{dialogue}
    
    \speak{Question} A candle melts by 2 centimetres every hour that it burns. How many centimetres shorter will a candle be after burning from 1:00 PM to 5:00 PM?

    \speak{Answer} 8 centimeters

    \speak{Initial} The candle will have burned for 5-1 = 4 hours. After 4 hours, the candle will have melted 4 * 2 = 8. Thus, the candle will be 8 - 2 = 6 centimeters shorter after burning from 1:00 PM to 5:00 PM.

    \speak{Looped} Burning from 1:00 PM to 5:00 PM is a total of 5 - 1 = 4 hours. The candle will melt 4 * 2 =8 centimeters shorter.
    
\end{dialogue}

In some cases, even though it didn't get the answer right, we could still see a logical improvement, as shown below.

\begin{dialogue}

    \speak{Question} A new program had 60 downloads in the first month. The number of downloads in the second month was three times as many as those in the first month but then reduced by 30\% in the third month. How many downloads did the program have in total over the three months?
    
    \speak{Answer} 366
    
    \speak{Initial} The number of downloads in the second month was 3*60=180 downloads. The number of downloads in the third month was 180*(100/100)=180 downloads. The total number of downloads over the three months is 60+180+180=420 downloads.
    
    \speak{Looped} The number of downloads in the second month was 3*60= 180. In the third month, the number of downloads was 180*.7=126. In total, the number of downloads in the three months was 126+180+60=426

\end{dialogue}

\section{Conclusion}
\label{conclusion}
We can see the performance of our network close to a fine-tuned LLM with equation-based context. Using our "Indusction Learning" approach we can see the results vary from 33\% of a fine-tuned model to 26\% for the initial network, finally improving up to 50.29\% for the distributed network. The key feature discriminating our network is the logic improvement of GPs. The GP improve their logic as we supply them with hints. In the future, using a large language model relying on reinforcement learning to tune the model, we will strive to improve the logic of the upper LLM as a standalone unit. We can use the trained LLM with higher logic to solve problems. This can be applied to various fields where the training of the initial model can be made more precise using these types of networks. 

Compared to state-of-the-art of the appraich, our model surpasses the Phi 1.5 baseline, which achieved a 40.2\% reasoning output using a solver, compared to our 50.29\%. Despite TinyGSM's impressive increase from 44.6\% to 68.2\%, our collaborative GP and EQ approach demonstrates competitive performance with a 25.29\% increase, showcasing advancements in mathematical reasoning capabilities.

\bibliography{main}
\bibliographystyle{icml2024}

\end{document}